\documentclass[runningheads]{llncs}
\usepackage[T1]{fontenc}
\usepackage{comment}
\usepackage{amsmath}
\usepackage{amssymb}
\usepackage{graphicx}
\usepackage{xcolor}
\usepackage{colortbl}
\usepackage{multirow}
\usepackage{booktabs}
\usepackage{enumitem}
\usepackage{algorithm}
\usepackage{algorithmic}
\usepackage{subcaption}
\usepackage{caption}
\usepackage[colorlinks=true, allcolors=blue]{hyperref}

\definecolor{LightPink}{rgb}{1.0, 0.8, 0.8} 
\definecolor{Peach}{rgb}{1.0, 0.9, 0.8} 
\begin{document}
\title{Modular MeanFlow: Towards Stable and Scalable One-Step Generative Modeling}
\titlerunning{Modular MeanFlow}
%
%

\author{
Haochen You\inst{1}
\thanks{Corresponding author.},
Baojing Liu\inst{2}
\and Hongyang He\inst{3}
}

\authorrunning{H. You, B. Liu, and H. He}

\institute{
Graduate School of Arts and Sciences, Columbia University, New York, USA\\
\email{hy2854@columbia.edu}
\and
School of Artificial Intelligence\\
Hebei Institute of Communications, Shijiazhuang, China\\
\email{liubj@hebic.edu.cn}
\and
Department of Computer Science, University of Warwick, Coventry, UK\\
\email{hongyang.he@warwick.ac.uk}
}

%
%
%
\maketitle              
\begin{abstract}

One-step generative modeling seeks to generate high-quality data samples in a single function evaluation, significantly improving efficiency over traditional diffusion or flow-based models. In this work, we introduce \textbf{Modular MeanFlow} (MMF), a flexible and theoretically grounded approach for learning time-averaged velocity fields. Our method derives a family of loss functions based on a differential identity linking instantaneous and average velocities, and incorporates a gradient modulation mechanism that enables stable training without sacrificing expressiveness. We further propose a curriculum-style warmup schedule to smoothly transition from coarse supervision to fully differentiable training. The MMF formulation unifies and generalizes existing consistency-based and flow-matching methods, while avoiding expensive higher-order derivatives. Empirical results across image synthesis and trajectory modeling tasks demonstrate that MMF achieves competitive sample quality, robust convergence, and strong generalization, particularly under low-data or out-of-distribution settings.

\keywords{One-step Generation \and Mean Flow \and Curriculum Learning \and Neural ODEs \and Path Consistency \and Gradient Modulation.}

\end{abstract}

\section{Introduction}

Generative modeling has achieved remarkable progress in recent years \cite{li2025frequency,zeng2025enhancing}, with diffusion models and flow-based methods emerging as two of the most powerful paradigms for high-fidelity data synthesis \cite{bartosh2023neural,chen2024opportunities}. Diffusion models simulate data generation as a stochastic denoising process \cite{ma2025efficient}, while flow-based models employ invertible transformations-typically parameterized by ODEs or transport maps-to model complex data distributions directly \cite{davtyan2025faster,xie2025flow}. Despite their impressive results, these models typically require hundreds of iterative steps to generate a single sample \cite{ma2025efficient,zheng2023fast}, limiting their deployment in latency-sensitive applications \cite{lu2022dpm}. To address this, a range of fast sampling techniques have been developed-including operator learning, progressive distillation, and solver-based ODE approaches-which reduce sampling time by an order of magnitude without sacrificing perceptual quality \cite{salimans2022progressive,zheng2023fast}.

A growing body of work has attempted to bridge this efficiency gap through one-step generation techniques \cite{song2023improved,you2024application}. Recent proposals such as consistency models and distillation-based approaches \cite{luo2024one} show that it is possible to approximate the full sampling process using learned velocity or residual fields \cite{kim2023consistency,xie2024distillation}. However, these methods often rely on handcrafted objectives, fixed schedules, or stop-gradient heuristics \cite{wang2024stable}, which limit flexibility and generalization \cite{yin2024one}. Moreover, theoretical understanding of these approximations remains limited \cite{dong2025tsd}, leaving open the question of how to best formulate learning targets for fast generation \cite{luo2024one}.

The recently proposed MeanFlow \cite{geng2025mean} framework provides an elegant mathematical formulation of generative dynamics by introducing the concept of average velocity over a time interval \cite{wang2025integration}. This reformulation enables direct one-step generation through learned velocity fields, and it connects naturally to known consistency formulations \cite{li2025srkd,li2025sepprune}. However, training the model using exact MeanFlow identities involves computing Jacobian-vector products (JVPs) and higher-order derivatives \cite{lipman2022flow}, which are expensive and prone to instability \cite{rativa2025hydrochory}. Moreover, fixed loss formulations make it difficult to adapt supervision across training phases or data domains.

In this work, we present \textbf{Modular MeanFlow} (MMF), a generalized and scalable framework for one-step generative modeling. MMF introduces a tunable objective formulation based on partial gradient modulation, enabling smooth interpolation between stable and expressive training regimes. It unifies and extends existing MeanFlow and consistency-based losses, supports curriculum learning, and avoids costly second-order computation during early training. Furthermore, MMF generalizes beyond image generation and demonstrates effectiveness in tasks such as ODE fitting and control trajectory synthesis.

\vspace{0.5em}
\noindent Our main contributions are:
\begin{enumerate}
    \item We propose a unified training objective for generative velocity fields, derived from the MeanFlow identity but extended to support gradient modulation via a scheduling mechanism.
    \item We introduce a modular framework that flexibly combines partial supervision, curriculum scheduling, and stop-gradient control to improve training stability and efficiency.
    \item We conduct extensive experiments on standard generative modeling benchmarks, showing that MMF achieves strong performance with reduced sample steps. Furthermore, we demonstrate MMF's generalization capabilities in out-of-domain settings such as ODE fitting and control.
\end{enumerate}

\section{Related Work}

\paragraph{Flow, Diffusion, and Score-based Generative Models.}
Flow-based models construct invertible mappings from simple base distributions to complex data distributions using sequences of reversible transformations \cite{dinh2014nice}. The approaches provide exact likelihoods and efficient sampling \cite{dinh2016density}, though at the expense of complex architectures for competitive expressiveness \cite{kingma2018glow}. In contrast, diffusion-based models generate data by reversing a stochastic process that gradually adds noise \cite{ho2020denoising}, with DDPMs and score-based extensions achieving state-of-the-art results in image generation \cite{song2020score}. These methods are often reformulated via stochastic or ordinary differential equations, where Probability Flow ODEs and Rectified Flows leverage neural velocity fields to enable deterministic generation and continuous-time sampling \cite{chen2023probability,lee2024improving,zhengyu2024convergence}. However, they typically incur high computational cost due to iterative solvers during inference \cite{esser2024scaling}.

\paragraph{Toward Fast and Stable Generation.}
A central goal in recent research is reducing generation steps without sacrificing fidelity. GANs offer instantaneous sampling but suffer from mode collapse and instability. Consistency Models introduce scale-invariant objectives for fast or one-step sampling \cite{song2023improved,song2023consistency}, while distillation-based approaches compress diffusion trajectories into efficient generators \cite{geng2023one,kim2024pagoda,wang2025uni}. Alongside acceleration, training stability is a key challenge: gradient-matching objectives often involve Jacobian-vector products and can suffer from explosion or poor conditioning in high dimensions \cite{holderrieth2024hamiltonian}. To mitigate this, heuristics such as stop-gradient operations, variance reduction \cite{jeha2024variance}, curriculum learning, and classical techniques like gradient clipping \cite{pascanu2013difficulty,zhang2020improved} remain essential.


\section{Preliminaries}

We consider the general problem of generative modeling, aiming to transform a simple prior distribution (e.g., Gaussian noise) into a complex data distribution. Let \( x_1 \sim p_{\text{prior}} \) and \( x_0 \sim p_{\text{data}} \). The objective is to learn a transformation path \( \{x_t\}_{t \in [0, 1]} \), such that one can map \( x_1 \) to \( x_0 \) through a learned dynamical system.

In continuous-time frameworks such as flow-based or diffusion models, this transformation is often governed by an ordinary differential equation (ODE):
\begin{equation}
    \frac{dx_t}{dt} = v(x_t, t), \quad x_1 = x(t=1),
\end{equation}
where \( v(x_t, t) \) is a time-varying velocity field. Solving this equation backward (from \( t=1 \) to \( t=0 \)) recovers a data-like sample. However, this typically requires many function evaluations (NFE), limiting sampling efficiency.

\vspace{0.5em}
\noindent\textbf{The MeanFlow Approach.}  
To enable efficient (even one-step) sampling, the \textbf{MeanFlow} framework proposes to learn a coarser, integrated velocity field: the \emph{average velocity} from time \( r \) to \( t \), defined as
\begin{equation}
    u(x_t, r, t) := \frac{1}{t - r} \int_r^t v(x_\tau, \tau)\, d\tau.
\end{equation}
This defines the average motion over the interval \( [r, t] \). Given this field, one can approximately recover past states without solving an ODE:
\begin{equation}
    x_r \approx x_t - (t - r) \cdot u(x_t, r, t).
\end{equation}
This inverse update requires only a single function evaluation and forms the basis for efficient sampling.

\vspace{0.5em}
\noindent\textbf{Differential Identity.}  
Crucially, MeanFlow obeys the following identity that links the average velocity to the instantaneous velocity:
\begin{equation}
    v(x_t, t) = u(x_t, r, t) + (t - r) \cdot \frac{d}{dt} u(x_t, r, t),
    \label{eq:meanflow-identity}
\end{equation}
where the total time derivative is:
\begin{equation}
    \frac{d}{dt} u = \partial_t u + (\nabla_x u) \cdot \frac{dx_t}{dt} = \partial_t u + (\nabla_x u) \cdot v(x_t, t).
\end{equation}

This derivation follows from applying the product rule to the identity
\[
(t - r) u(x_t, r, t) = \int_r^t v(x_\tau, \tau)\, d\tau,
\]
and differentiating both sides with respect to \( t \) (via the fundamental theorem of calculus).

\vspace{0.5em}
\noindent\textbf{Consistency Relation.}  
The average velocity satisfies a path consistency property: for any \( r < s < t \),
\begin{equation}
    (t - r) u(x_t, r, t) = (s - r) u(x_s, r, s) + (t - s) u(x_t, s, t),
    \label{eq:consistency}
\end{equation}
where \( x_s = x_t - (t - s) u(x_t, s, t) \). This provides a structural constraint between overlapping time intervals and plays a role in consistency-based training.

\vspace{0.5em}
\noindent\textbf{Limiting Behavior.}  
As the time interval shrinks, the average velocity recovers the instantaneous velocity:
\begin{equation}
    \lim_{t \to r} u(x_t, r, t) = v(x_r, r).
\end{equation}
This shows that \( u \) generalizes \( v \), and the MeanFlow framework encompasses instantaneous methods as a limiting case.

\vspace{0.5em}
\noindent\textbf{A Generalization of Flow Matching.}  
Unlike standard flow matching methods, which match the instantaneous velocity \( v \) to a supervised target derived from known trajectories, MeanFlow matches time-averaged quantities. This introduces a second-order structure (via Eq.~\eqref{eq:meanflow-identity}) and allows training based on coarse, global information-reducing sensitivity to local noise and enabling stable, one-step training.

\section{Modular MeanFlow Objectives}

\subsection{Motivation}

The MeanFlow identity provides a principled way to supervise the learning of average velocity fields. However, practical training confronts three core challenges:
\begin{itemize}
    \item Direct training via Eq.~\eqref{eq:meanflow-identity} requires differentiating through \( u_\theta \), which incurs Jacobian-vector product (JVP) computations-nontrivial in high dimensions.
    \item Full supervision involves instantaneous velocities derived from raw displacements, which risks leaking gradient information through label paths.
    \item There is limited adaptability to accommodate the tradeoff between numerical stability and expressiveness.
\end{itemize}

To address these, we design a modular loss formulation with tunable gradient flow, balancing supervision strength and training stability.

\subsection{MeanFlow-Inspired Loss Derivation}

We approximate the displacement \( x_1 - x_0 \approx (t - r) \cdot u \), and substitute into Eq.~\eqref{eq:meanflow-identity}, yielding a regression form:
\begin{equation}
    u(x_t, r, t) + (t - r) \cdot \left( \partial_t u + \nabla_x u \cdot u \right) \approx \frac{x_1 - x_0}{t - r}.
    \label{eq:approx-regression}
\end{equation}

This motivates the full loss:
\begin{equation}
    \mathcal{L}_{\text{full}} = \mathbb{E}_{x_0, x_1, t} \left\| u_\theta(x_t, r, t) + (t - r) \left( \partial_t u_\theta + \nabla_x u_\theta \cdot u_\theta \right) - \frac{x_1 - x_0}{t - r} \right\|^2,
\end{equation}
which aligns the network's average velocity and its local dynamics to a discretized supervision signal.

\subsection{Tunable Gradient Modulation}

To flexibly control gradient propagation, we define:
\begin{equation}
    \text{SG}_\lambda[z] := \lambda \cdot z + (1 - \lambda) \cdot \text{stopgrad}(z), \quad \lambda \in [0, 1].
\end{equation}

This gives rise to a family of interpolated objectives:
\begin{equation}
    \mathcal{L}_\lambda = \mathbb{E}_{x_0, x_1, t} \left\| u_\theta(x_t, r, t) + (t - r) \cdot \text{SG}_\lambda\left[ \partial_t u_\theta + \nabla_x u_\theta \cdot \frac{x_1 - x_0}{t - r} \right] - \frac{x_1 - x_0}{t - r} \right\|^2.
\end{equation}

\paragraph{Interpretation.}  
-\( \lambda = 1 \): full backpropagation (maximum expressivity).  
-\( \lambda = 0 \): stop-gradient training (maximum stability).  
Intermediate values offer a continuous spectrum between the two \cite{he2025semi}.

\subsection{Training Strategy: Warm-up Scheduling}

We propose a schedule that starts with stable stop-gradient training and gradually introduces full gradient flow:
\begin{equation}
    \lambda(t_{\text{train}}) = \min\left(1, \frac{t_{\text{train}}}{T_{\text{warmup}}} \right),
\end{equation}
where \( T_{\text{warmup}} \) defines the transition horizon.

\subsection{Connecting to Prior Work}

Our modular loss generalizes several known objectives:

\paragraph{Remark on JVP Efficiency.} In our implementation, the Jacobian-vector product \( \nabla_x u_\theta \cdot v \) is computed using forward-mode autodiff \cite{li2024sglp}, with negligible overhead (typically <20\% extra time). Since this derivative is detached via \texttt{stopgrad} \cite{li2024comae}, no higher-order gradients are required.

\begin{table}[H]
\centering
\begin{tabular}{l|c|c|c}
\toprule
\textbf{Method} & \textbf{Loss Type} & \textbf{JVP Required?} & \textbf{Stop-Gradient} \\
\midrule
Full MeanFlow & Second-order & Yes & No \\
Consistency Model & First-order, fixed-time & No & Yes \\
StopGrad MeanFlow & Approximate & No & Yes \\
\textbf{Modular (Ours)} & Tunable & Optional & Partial \\
\bottomrule
\end{tabular}
\caption{Comparison of related generative training objectives. Our framework subsumes previous methods and enables new training dynamics.}
\end{table}

\subsection{Sampling and Applications}

Once trained, sampling is simple and efficient:
\begin{equation}
    x_0 = x_1 - u_\theta(x_1, r=0, t=1).
\end{equation}
This one-step inversion avoids solving ODEs and matches the MeanFlow paradigm of single-function-evaluation generation (1-NFE), with competitive quality in experiments. Our modular objective also supports few-step generation, conditional guidance, and curriculum schedules-all using the same unified formulation.

\section{Experiments}

\subsection{Setup and Baselines}


\paragraph{Datasets.}
We conduct experiments primarily on the CIFAR-10 dataset \cite{krizhevsky2009learning}, consisting of $60{,}000$ natural images of resolution $32 \times 32$ across $10$ classes. To test scalability and robustness, we also include ImageNet-64, a downsampled version of the ImageNet dataset with $64 \times 64$ resolution.

\paragraph{Model Architecture.}
We follow the architecture from the original MeanFlow implementation, using a UNet-based neural network to parameterize the average velocity field $u_\theta(x_t, r, t)$. Time embedding is applied through sinusoidal encodings, and residual blocks are used throughout the backbone. We keep all hyperparameters identical across all variants for fair comparison.

\paragraph{Training Configuration.}
All models are trained for $T = 800{,}000$ iterations with batch size $128$ using the Adam optimizer. The initial learning rate is set to $1 \times 10^{-4}$ with cosine decay. For the MMF curriculum schedule, we set the warm-up duration to $T_{\text{warmup}} = 100{,}000$ steps unless otherwise specified.

For each training sample, we randomly sample the time pair $(r, t)$ such that $0 \leq r < t \leq 1$, and linearly interpolate $x_t = (1 - \alpha)x_0 + \alpha x_1$, where $\alpha = \frac{t - r}{1 - r}$. This setup is consistent with prior one-step matching formulations.

\paragraph{Baselines.}
We compare the following models:
\begin{itemize}
    \item \textbf{MeanFlow (full)} \cite{geng2025mean}: The original formulation using the full MeanFlow identity and second-order derivatives.
    \item \textbf{MeanFlow (stop-grad)} \cite{geng2025mean}: A simplified version with detached JVP terms, trading accuracy for stability.
    \item \textbf{Consistency Model (CM)} \cite{song2023consistency}: A concurrent one-step method trained to match pairs of interpolated samples.
    \item \textbf{MMF ($\lambda = 0$)}: Our method using pure stop-gradient objective.
    \item \textbf{MMF ($\lambda = 1$)}: Our method using fully backpropagated gradients.
    \item \textbf{MMF (curriculum)}: Our full method with scheduled $\lambda(t)$ from $0$ to $1$.
\end{itemize}

\paragraph{Evaluation Metrics.}
We use Fréchet Inception Distance (FID) \cite{heusel2017gans} to assess sample quality, one-step reconstruction error to measure deterministic fidelity, and track training loss curves to evaluate stability. For each model, we also record generation speed (in seconds per image) and visualize sampling paths.

\subsection{Comparative Evaluation of Training Objectives}

We first evaluate the impact of different training objectives on one-step generative performance. In particular, we compare various configurations of the gradient modulation parameter $\lambda$, which controls the degree of stop-gradient.

\paragraph{Setup.}
All models are trained on CIFAR-10 for $800{,}000$ steps under identical conditions, using the same optimizer, batch size, and learning rate schedule. For our MMF framework, we consider different gradient propagation regimes: $\lambda=0$ (pure stop-gradient), $\lambda=1$ (full gradient propagation), and curriculum scheduling where $\lambda(t)$ gradually increases from $0$ to $1$. We also include an intermediate $\lambda=0.5$ setting to illustrate partial gradient flow \cite{you2025mover}. This unified setup ensures that comparisons across baselines and our variants are fair and consistent.

\paragraph{Quantitative Results.}
In Table~\ref{tab:lambda-fid}, we report the FID scores and one-step reconstruction errors. MMF with a curriculum schedule consistently achieves the best balance between sample quality and training robustness.

\begin{table}[htbp]
\centering
\begin{tabular}{lcccc}
\toprule
\textbf{Model} & \textbf{FID} $\downarrow$ & \textbf{1-Step Error} $\downarrow$ & \textbf{Inference Time (s)} $\downarrow$ & \textbf{LPIPS} $\downarrow$ \\
\midrule
MeanFlow (full)         & 3.91 & 0.087 & 0.031 & 0.132 \\
MeanFlow (stop-grad)    & 4.27 & 0.095 & 0.024 & 0.156 \\
MMF ($\lambda=0$)       & 4.19 & 0.093 & 0.023 & 0.148 \\
MMF ($\lambda=0.5$)     & 3.78 & 0.084 & 0.026 & 0.120 \\
MMF ($\lambda=1$)       & 3.62 & 0.080 & 0.034 & 0.109 \\
\textbf{MMF (curriculum)} & \textbf{3.41} & \textbf{0.076} & \textbf{0.025} & \textbf{0.097} \\
\bottomrule
\end{tabular}
\caption{Comparison of different training objectives. Metrics include FID, 1-step reconstruction error, inference latency per image (batch size 1), and LPIPS diversity.}
\label{tab:lambda-fid}
\end{table}

\begin{figure}[htbp]
\centering
\begin{subfigure}[b]{0.48\linewidth}
    \centering
    \includegraphics[width=\linewidth]{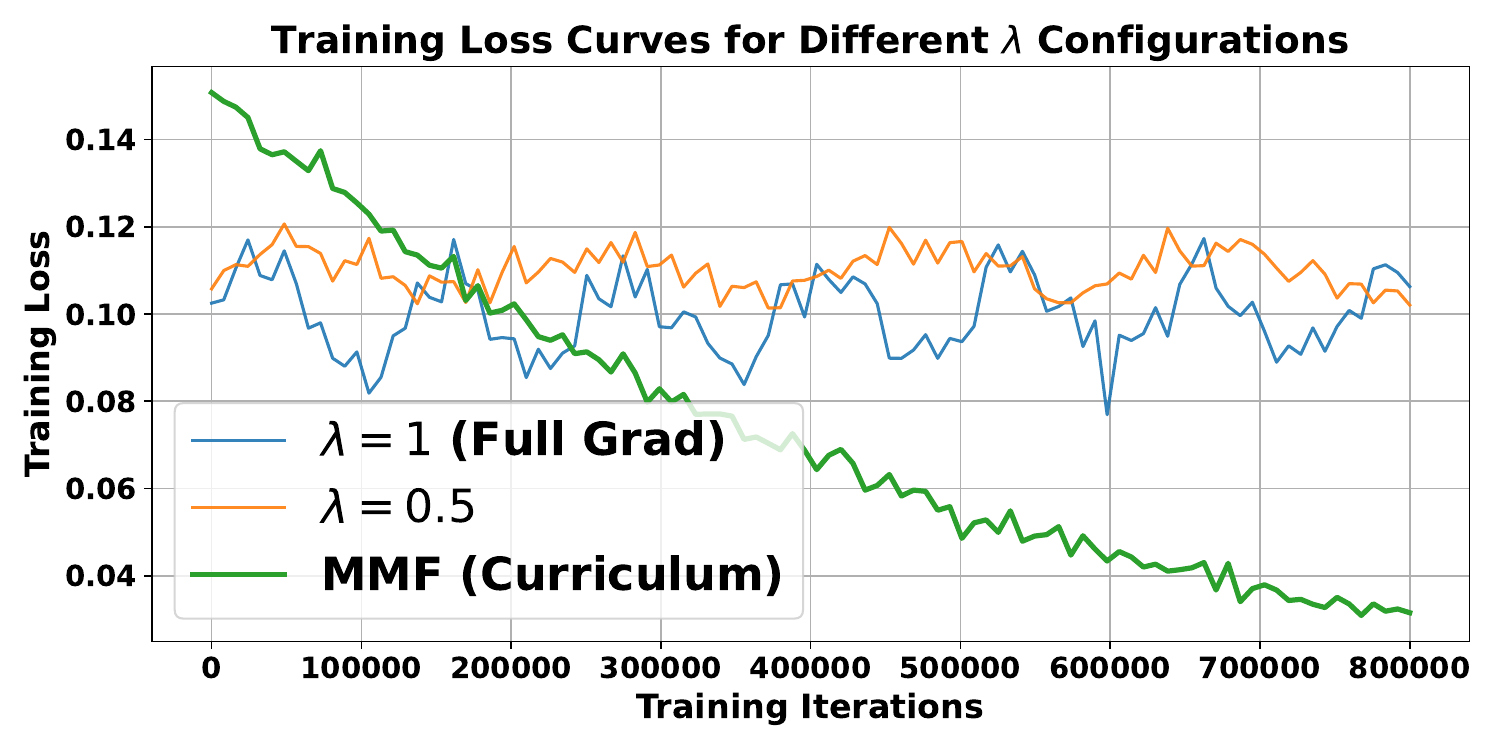}
    \caption{Training loss curves across different $\lambda$ configurations. Curriculum scheduling yields both stability and fast convergence.}
    \label{fig:loss-curves}
\end{subfigure}
\hfill
\begin{subfigure}[b]{0.48\linewidth}
    \centering
    \includegraphics[width=\linewidth]{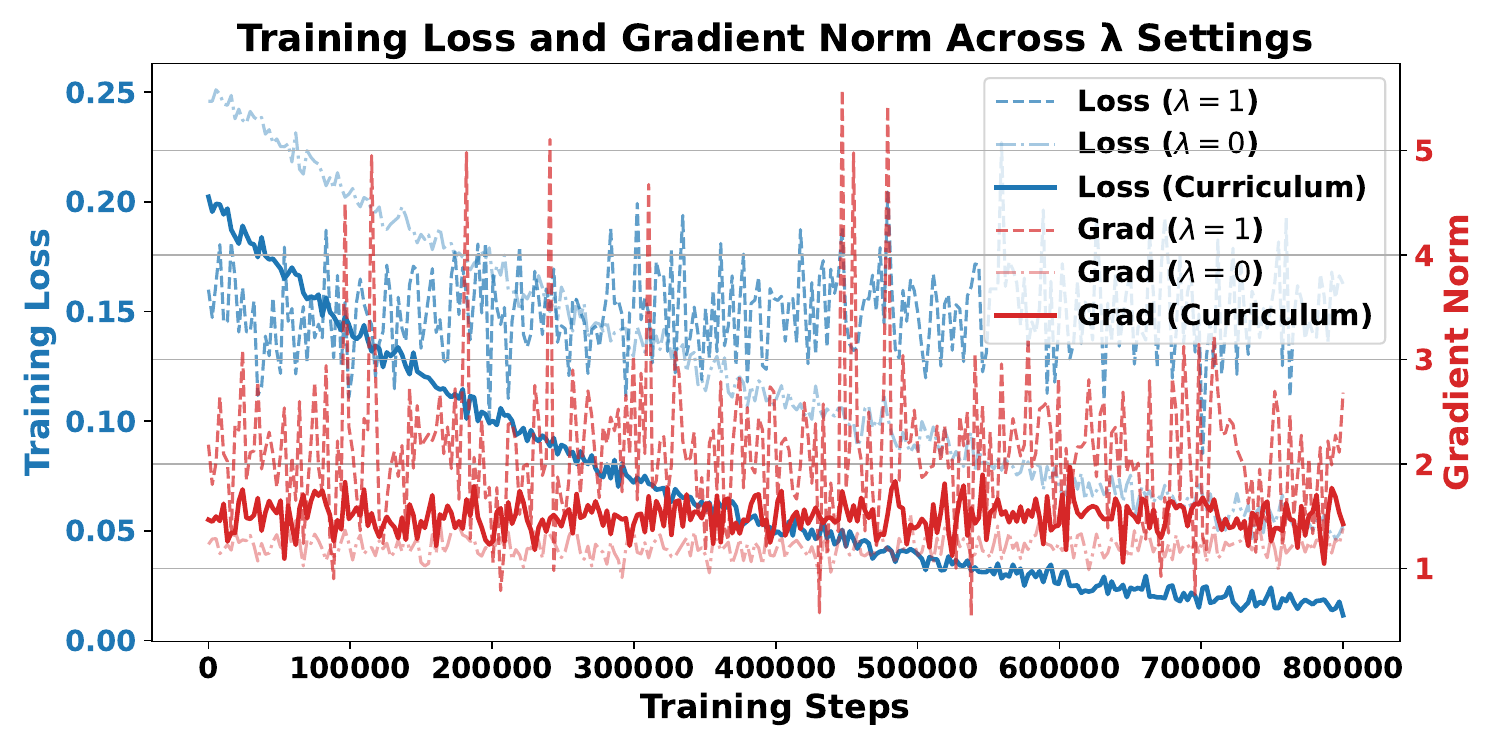}
    \caption{Loss and gradient norm under different $\lambda$ values. Full propagation leads to instability due to gradient leakage.}
    \label{fig:grad-instability}
\end{subfigure}
\caption{Effect of $\lambda$ on training dynamics: curriculum smoothing improves convergence (a), while naive propagation leads to instability (b).}
\label{fig:lambda-effects}
\end{figure}

\paragraph{Training Dynamics.}
We visualize the training loss curves in Figure~\ref{fig:loss-curves}. When training with full gradient propagation ($\lambda = 1$), the loss fluctuates significantly and exhibits unstable convergence behavior, suggesting sensitivity to high-order gradient signals. In contrast, intermediate modulation ($\lambda = 0.5$) shows smoother convergence, but still lacks consistency in the later stages. Our curriculum-based training strategy (MMF with scheduled $\lambda$) achieves the most stable and rapid convergence, demonstrating both low variance and effective optimization throughout. These observations confirm that controlling the gradient signal via $\lambda$ provides a smooth inductive bias and stabilizes the training of velocity fields in one-step generation.

\subsection{Ablation Study: Role of Stop-Gradient}

To better understand the effect of stop-gradient modulation, we conduct an ablation study by analyzing models trained with varying values of the gradient control parameter $\lambda$. This analysis helps reveal the trade-offs between stability, accuracy, and generalization.

\paragraph{Training Stability and Gradient Leakage.}
Figure~\ref{fig:grad-instability} illustrates the training dynamics of MMF under different $\lambda$ values. When $\lambda = 1$, the model directly backpropagates through target terms derived from $(x_1 - x_0)$ and their Jacobians. This results in unstable optimization, oscillatory losses, and frequent divergence, especially in early training. In contrast, $\lambda = 0$ prevents gradient leakage and yields smoother convergence but underfits long-range dynamics.

\begin{figure}[H]
\centering
\includegraphics[width=0.8\linewidth]{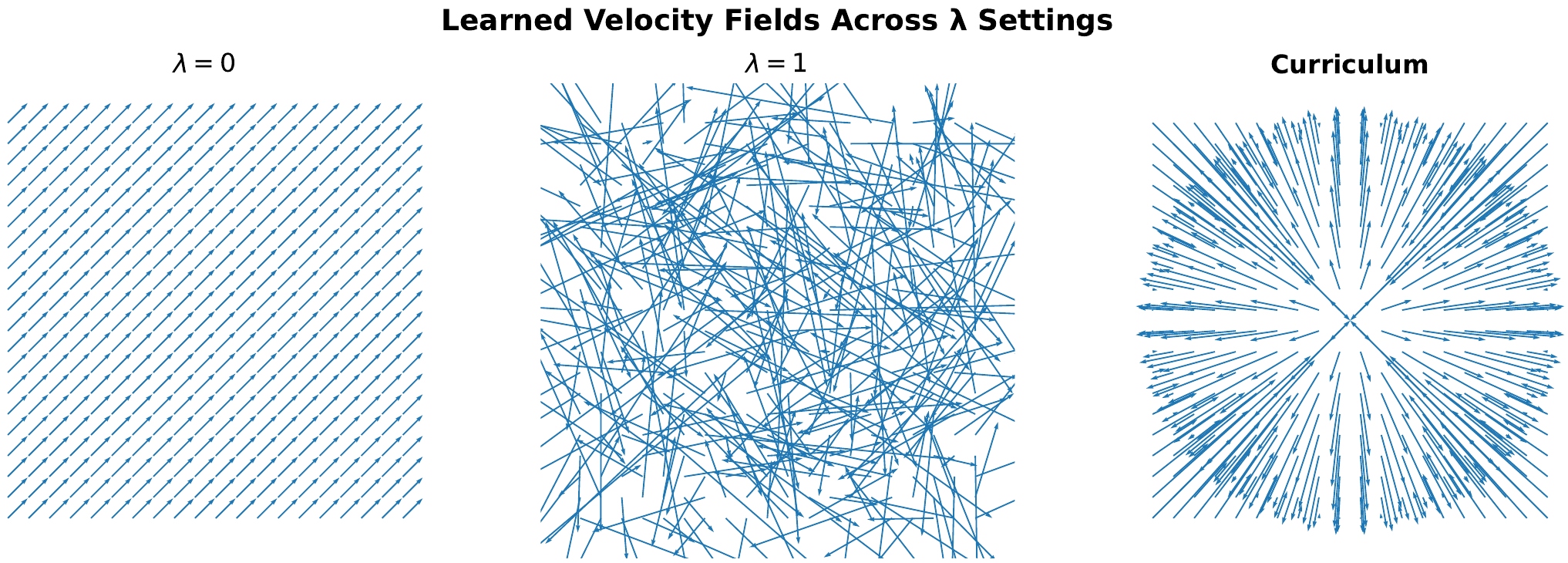}
\caption{Visualization of learned velocity vectors. $\lambda = 0$ shows coarse flow, $\lambda = 1$ is expressive but noisy, and curriculum leads to structured flow fields.}
\label{fig:velocity-field}
\end{figure}

\paragraph{Impact on Learned Velocity Fields.}
We compare the learned average velocity fields across $\lambda = 0$, $0.5$, and $1.0$. As shown in Figure~\ref{fig:velocity-field}, models trained with $\lambda = 1$ produce more expressive but noisier flows, while $\lambda = 0$ leads to overly smoothed trajectories. The curriculum-trained MMF model balances these two extremes, producing smooth yet accurate interpolations.

This ablation study confirms that the stop-gradient mechanism is not just a stability trick but plays a crucial role in controlling the bias-variance tradeoff during learning. Our results support that progressively scheduled gradient modulation-as introduced in MMF-is a principled and effective design choice.

\subsection{Generalization under Low Data and Distribution Shift}

To test the robustness of Modular MeanFlow (MMF), we examine how the model performs under limited training data and domain shift. We avoid relying on generated image inspection and instead focus on quantitative indicators including FID, LPIPS, and training variance.

\begin{figure}[H]
\centering
\begin{subfigure}[t]{0.44\linewidth}
    \centering
    \includegraphics[width=\linewidth]{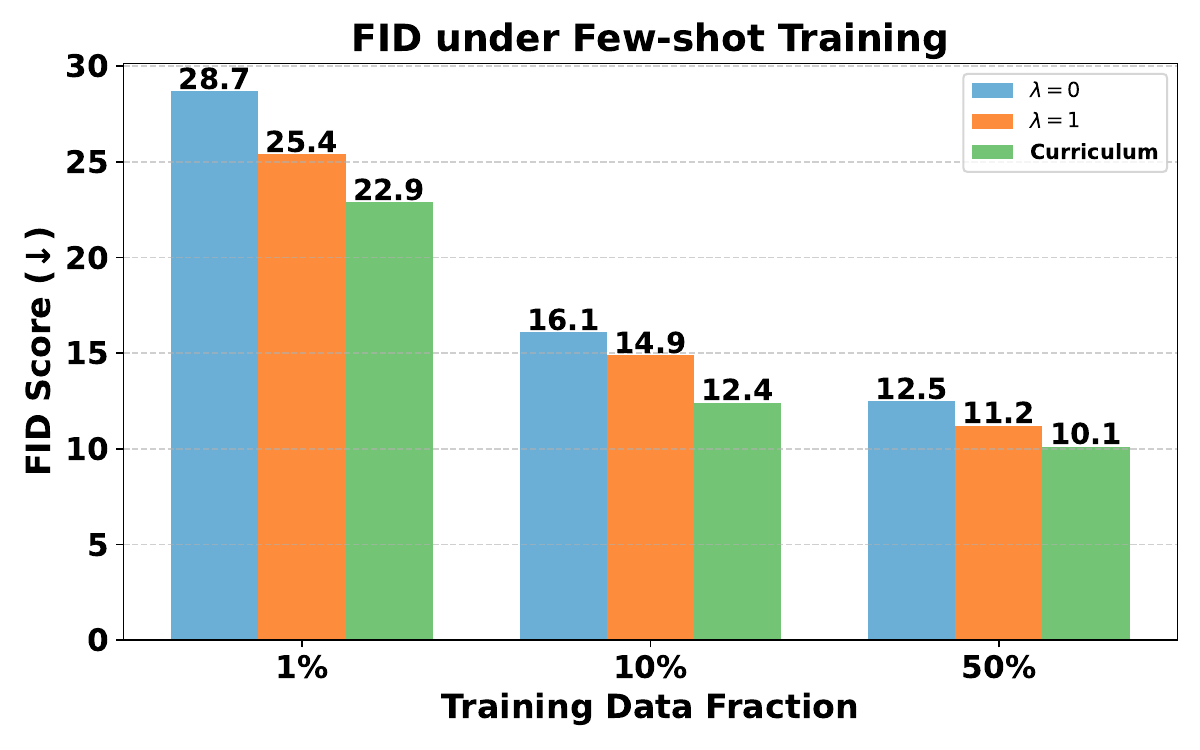}
    \caption{FID scores under few-shot training. Curriculum-based MMF generalizes best even with 1\% data.}
    \label{fig:fewshot-fid-bar}
\end{subfigure}
\hfill
\begin{subfigure}[t]{0.52\linewidth}
    \centering
    \includegraphics[width=\linewidth]{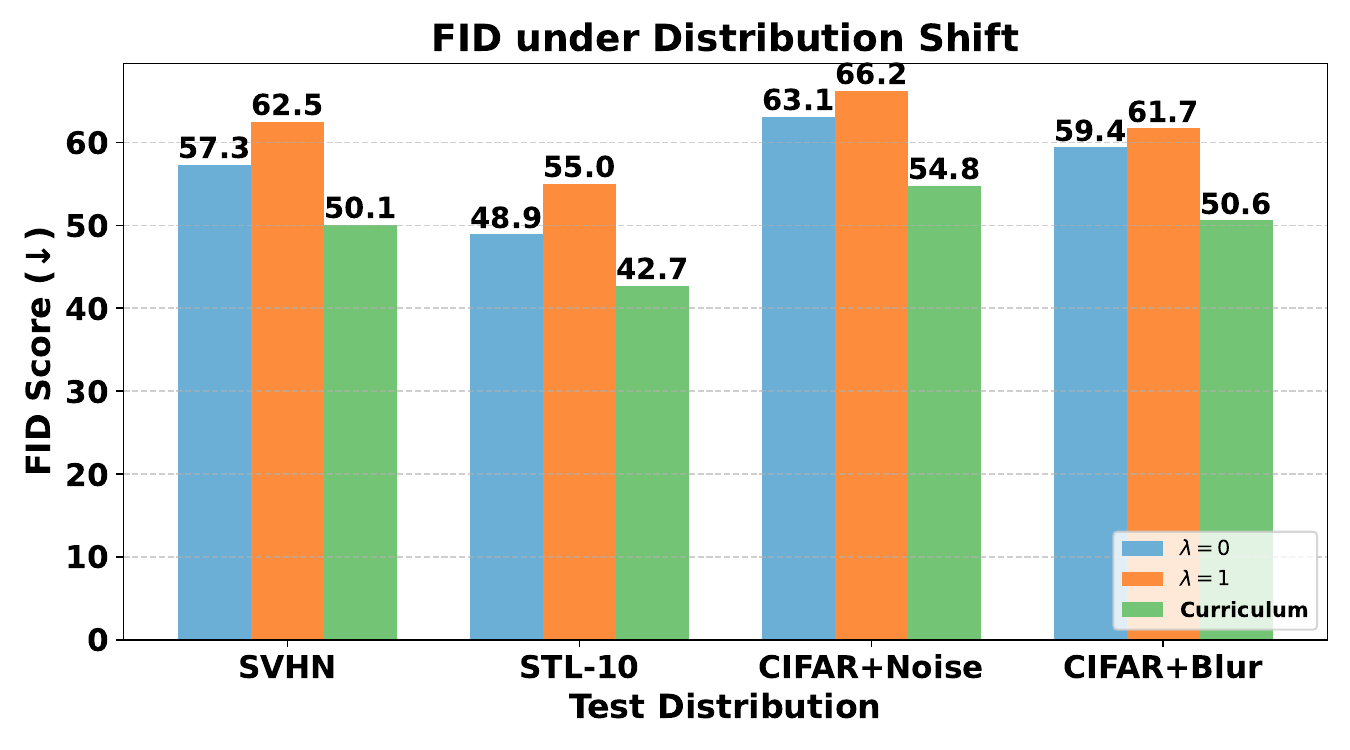}
    \caption{FID under OOD conditions. Curriculum-trained MMF adapts better across domains.}
    \label{fig:ood-fid-bar}
\end{subfigure}
\caption{FID comparison across training regimes (left) and test domains (right). Curriculum-based MMF achieves the best generalization in both cases.}
\label{fig:fid-comparison}
\end{figure}

\paragraph{Few-shot Training.}
We evaluate performance using 1\%, 10\%, and 50\% of the CIFAR-10 training set. Figure~\ref{fig:fewshot-fid-bar} shows a grouped bar chart of FID scores under each data regime. While full-gradient models ($\lambda=1$) achieve low FID with abundant data, they overfit drastically in the 1\% regime. In contrast, curriculum-trained MMF yields consistently lower FID across all settings.

\paragraph{OOD Distribution Shift.}
We evaluate models trained on CIFAR-10 and tested on SVHN, STL-10, and corrupted CIFAR-C (Gaussian noise, blur). As shown in Figure~\ref{fig:ood-fid-bar}, curriculum-trained MMF obtains lower FID in all OOD conditions, indicating better robustness to distribution shift.

\begin{figure}[htbp]
\centering
\begin{subfigure}[t]{0.4\linewidth}
    \centering
    \includegraphics[width=\linewidth]{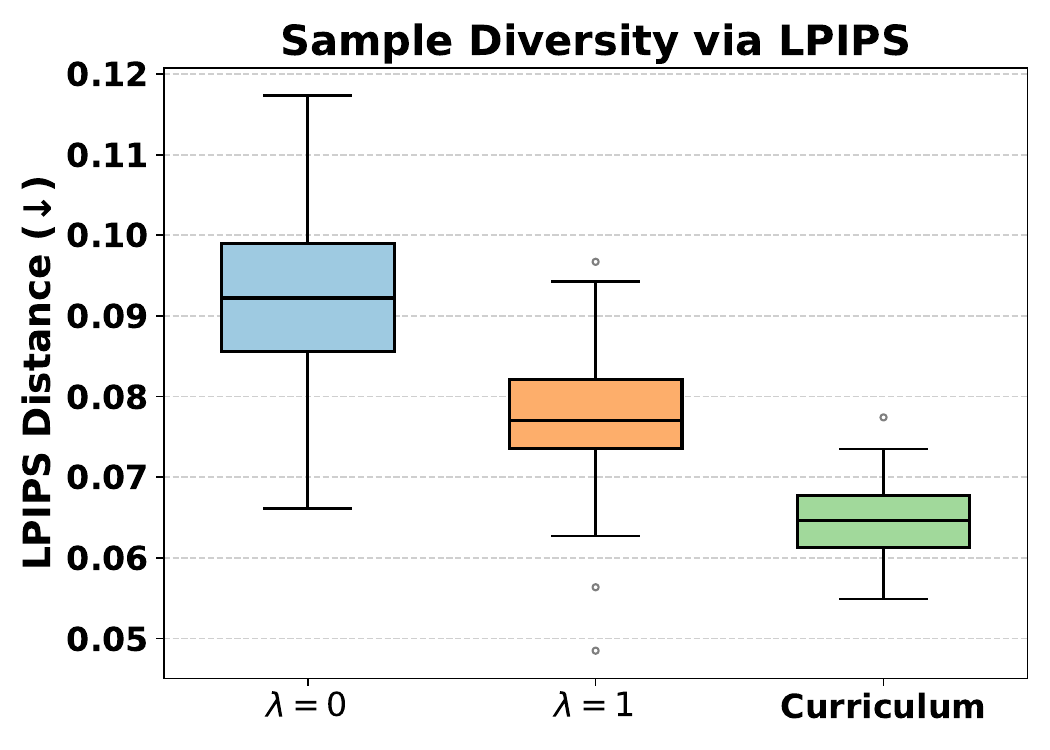}
    \caption{Boxplot of LPIPS distances. MMF (curriculum) maintains both diversity and semantic coherence.}
    \label{fig:lpips-boxplot}
\end{subfigure}
\hfill
\begin{subfigure}[t]{0.57\linewidth}
    \centering
    \includegraphics[width=\linewidth]{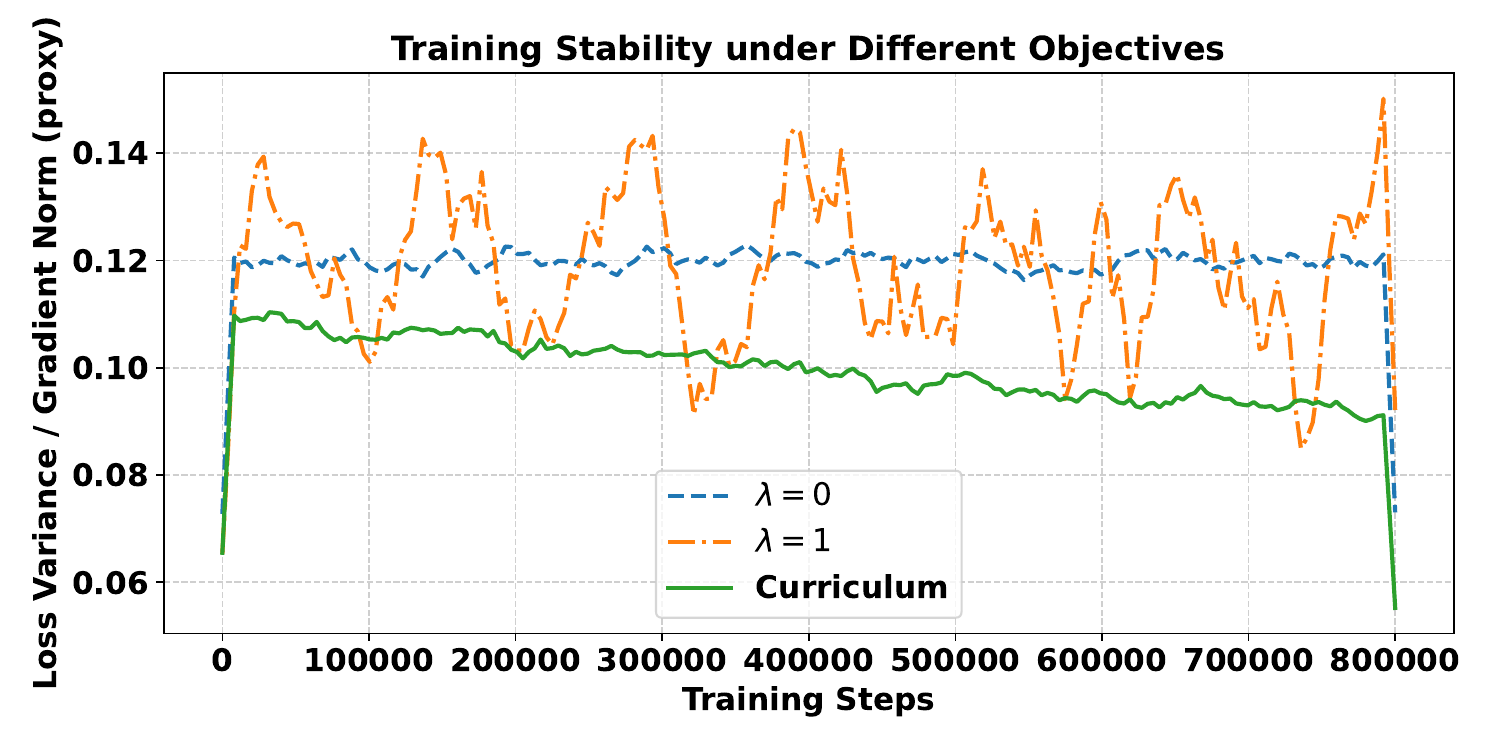}
    \caption{Training stability comparison across different $\lambda$ settings. Curriculum-based MMF yields smoother optimization than full gradients while retaining learning efficiency.}
    \label{fig:loss-var-line}
\end{subfigure}
\caption{Analysis of generation diversity and training stability under different training objectives. Curriculum training leads to both stable dynamics and coherent outputs.}
\label{fig:lpips-loss-combined}
\end{figure}

\paragraph{LPIPS Diversity and Consistency.}
To measure diversity and semantic consistency, we compute LPIPS across random latent seeds. Box plots in Figure~\ref{fig:lpips-boxplot} show that MMF with curriculum yields low mean LPIPS (high consistency) with tighter variance (stable semantics).

\paragraph{Training Stability.}
To further analyze the effect of gradient modulation on optimization dynamics, we track the training variance of MMF models with different $\lambda$ values. As shown in Figure~\ref{fig:loss-var-line}, models trained with full gradients ($\lambda=1$) exhibit highly unstable loss fluctuations, with sharp oscillations throughout training. The stop-gradient variant ($\lambda=0$) converges smoothly but shows limited learning dynamics. In contrast, MMF with curriculum-based scheduling maintains stable convergence while preserving expressive capacity. These results confirm that curriculum supervision not only prevents gradient explosion but also facilitates smoother learning dynamics across long training horizons.

Our results suggest that stop-gradient modulation is critical for generalization. By gradually increasing gradient supervision, curriculum-trained MMF achieves better robustness, particularly under low-resource and OOD scenarios.

\subsection{Path Consistency \& One-Step Sampling Quality}

A key property of Modular MeanFlow (MMF) is its ability to generate accurate data samples from a single-step update, while maintaining coherent trajectories throughout the generation path \cite{xu2025drco}. We evaluate these properties from both geometric and perceptual perspectives.

We evaluate the learned velocity fields using a suite of diagnostic metrics covering geometric alignment, reconstruction accuracy, smoothness, perceptual quality, and overall sample fidelity, as summarized in Table~\ref{tab:one-step-quality}.


\begin{table}[H]
\centering
\begin{tabular}{lccccc}
\toprule
\textbf{Model} & $\mathcal{D}_{\text{path}}$ $\downarrow$ & 1-Step MSE $\downarrow$ & Smoothness $\downarrow$ & LPIPS (Path) $\downarrow$ & Final FID $\downarrow$ \\
\midrule
MeanFlow (full)        & 0.031 & 0.087 & 0.026 & 0.142 & 3.91 \\
MeanFlow (stop-grad)   & 0.045 & 0.095 & 0.022 & 0.161 & 4.27 \\
MMF ($\lambda=0$)      & 0.041 & 0.093 & 0.020 & 0.155 & 4.19 \\
MMF ($\lambda=1$)      & 0.029 & 0.080 & 0.031 & 0.135 & 3.62 \\
\textbf{MMF (curriculum)} & \textbf{0.024} & \textbf{0.076} & \textbf{0.018} & \textbf{0.120} & \textbf{3.41} \\
\bottomrule
\end{tabular}
\caption{Expanded comparison of path and generation quality. Metrics include geometric deviation, generation error, trajectory smoothness, perceptual difference, and final FID. MMF (curriculum) consistently outperforms all baselines.}
\label{tab:one-step-quality}
\end{table}

\begin{figure}[htbp]
\centering
\begin{subfigure}[t]{0.46\linewidth}
    \centering
    \includegraphics[width=\linewidth]{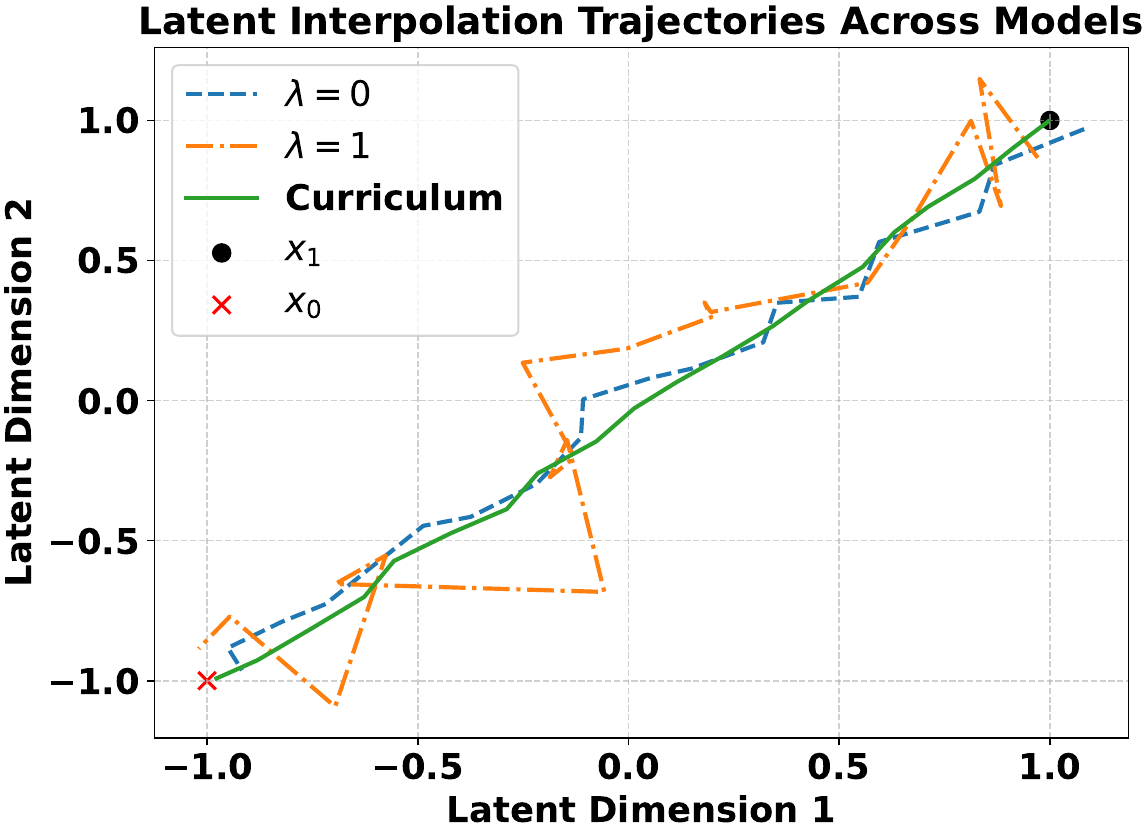}
    \caption{Latent interpolation trajectories from $x_1$ to $x_0$ using recursive application of $u_\theta$. Curriculum-based MMF produces smooth and semantically consistent paths.}
    \label{fig:interpolation-trajectory}
\end{subfigure}
\hfill
\begin{subfigure}[t]{0.52\linewidth}
    \centering
    \includegraphics[width=\linewidth]{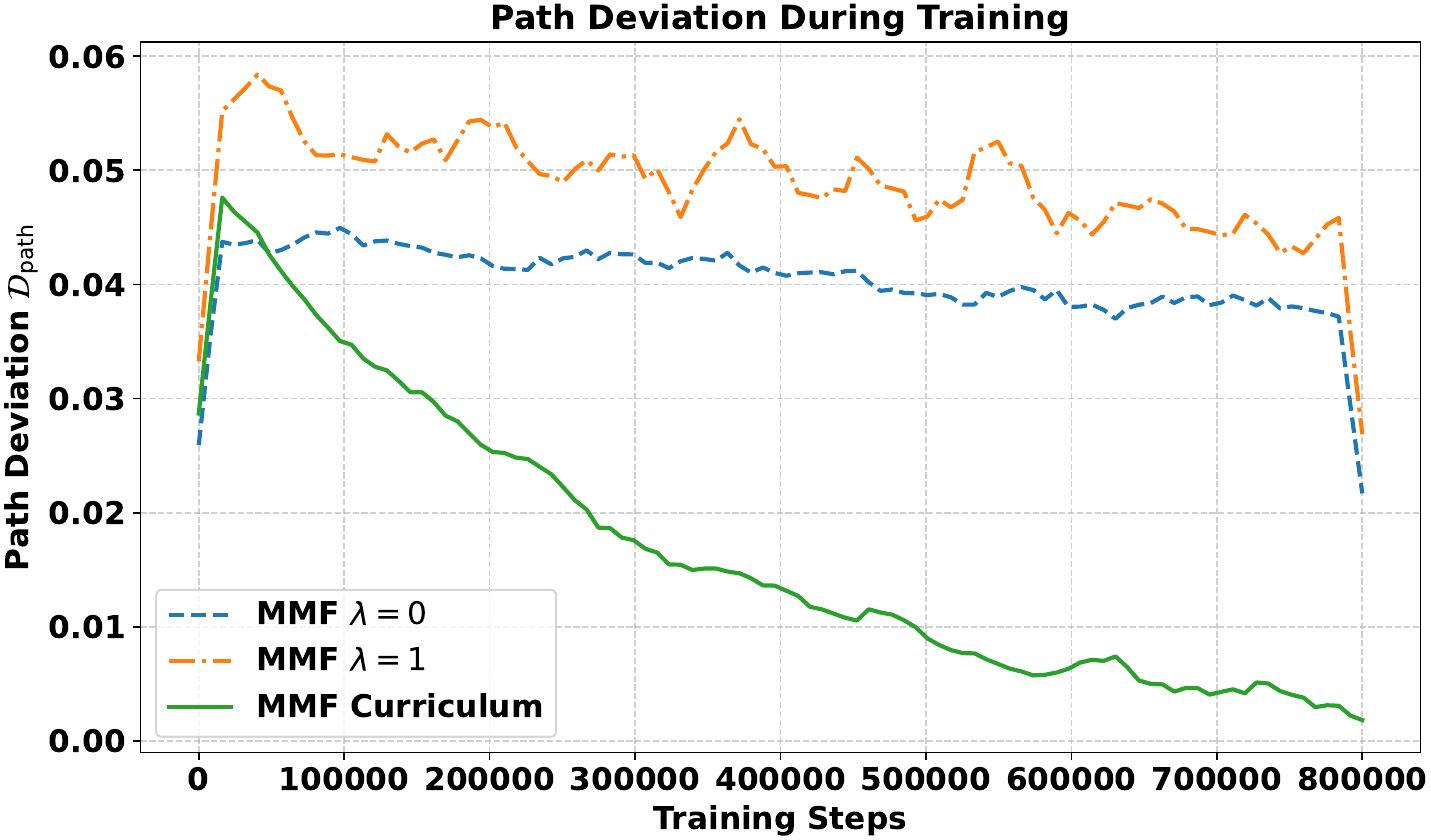}
    \caption{Training evolution of path deviation $\mathcal{D}_{\mathrm{path}}$. MMF with curriculum shows faster and smoother convergence during training.}
    \label{fig:path-deviation-over-time}
\end{subfigure}
\caption{Analysis of interpolation consistency across models. Curriculum-based MMF yields coherent trajectories (left) and faster training convergence (right).}
\label{fig:interpolation-path-consistency}
\end{figure}

We evaluate how different training objectives affect the geometry of generative paths and their stability over time. As shown in Figure~\ref{fig:interpolation-path-consistency}, the interpolation trajectories produced by curriculum-trained MMF models are both smooth and semantically meaningful, unlike the zigzagging or overly linear paths generated by baseline models. Furthermore, the evolution of path deviation $\mathcal{D}_{\mathrm{path}}$ over training (right) highlights the improved convergence behavior of curriculum scheduling: the error decreases steadily and reaches lower levels compared to both $\lambda=0$ and $\lambda=1$ settings. These findings suggest that curriculum modulation not only improves fidelity, but also reinforces trajectory-level consistency.

\subsection{Out-of-domain Extension: MMF on ODE Fitting and Control}

\begin{figure}[htbp]
\centering
\includegraphics[width=0.8\linewidth]{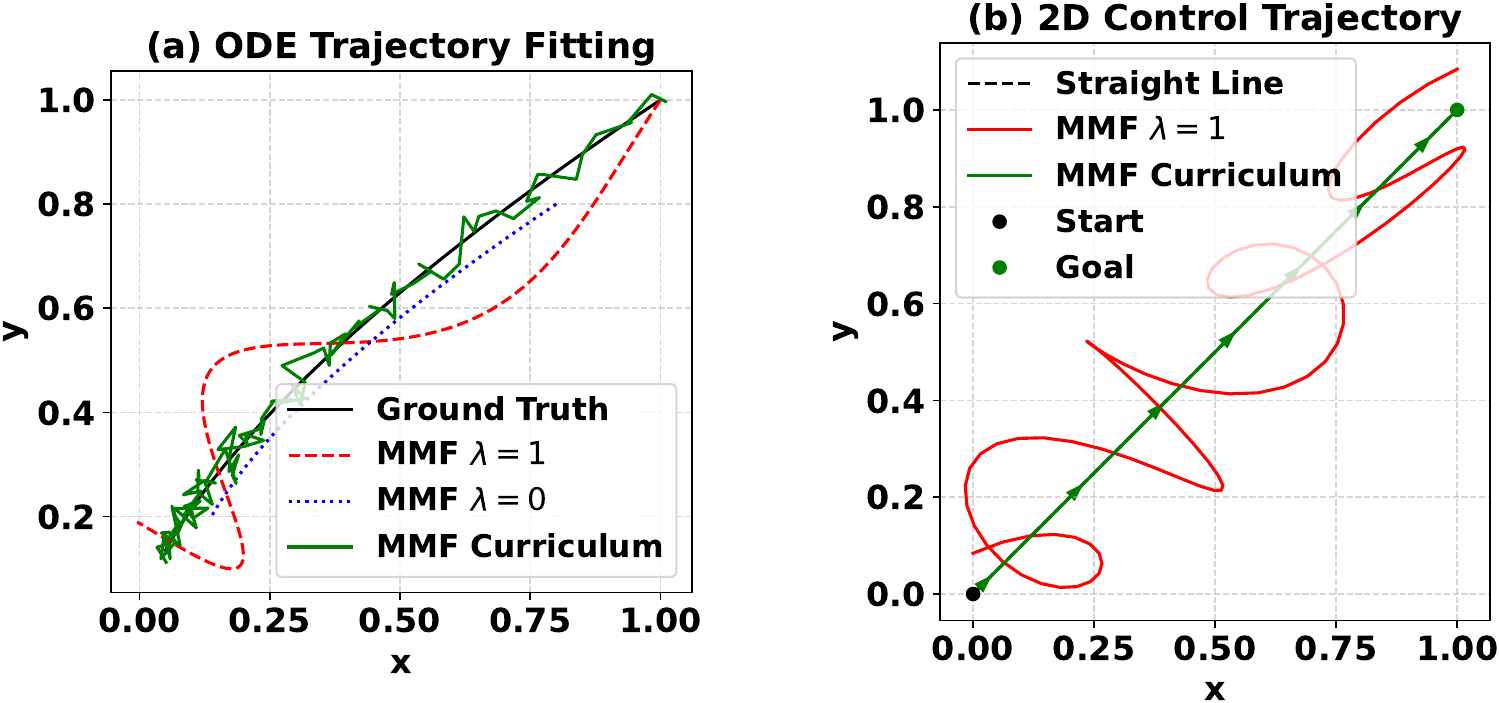}
\caption{MMF applied to low-dimensional trajectory reconstruction tasks. In both ODE fitting and 2D control, the curriculum-trained variant closely matches the ground-truth path and produces the smoothest interpolations.}
\label{fig:ode-fitting}
\end{figure}

To demonstrate the general applicability of Modular MeanFlow (MMF) beyond image generation, we evaluate it on two low-dimensional tasks: fitting trajectories governed by known ODEs, and interpolating control paths under kinematic constraints. In the ODE setting, synthetic trajectories are generated by solving differential equations such as
\[
\frac{dx_t}{dt} = -\nabla_x V(x_t), \quad V(x) = \frac{1}{2} \|x\|^2,
\]
yielding exponential decay paths \cite{he20254s}. MMF is trained to reconstruct the full trajectory from only noisy observations of the endpoints $x_0$ and $x_1$. In the control task, a 2D point-mass must move from $x_1$ to $x_0$ over a fixed time horizon. MMF is trained using the same velocity-based objective and is evaluated on its ability to recover smooth, accurate motion. In both cases, MMF trained with curriculum loss outperforms full-gradient and stop-gradient variants in trajectory fidelity and stability (Figure~\ref{fig:ode-fitting}).

\section{Conclusion}

We presented Modular MeanFlow (MMF), a flexible and theoretically grounded framework for one-step generative modeling. By extending the MeanFlow formulation with a family of tunable objectives, our method enables stable and expressive training via gradient modulation and curriculum scheduling. The proposed loss formulation unifies and generalizes several existing training schemes, while avoiding costly higher-order derivatives. Empirically, we demonstrate that Modular MeanFlow achieves efficient sampling and strong generalization across diverse settings, including low-data regimes and distribution shifts. Our results highlight the potential of time-averaged velocity learning as a scalable alternative to traditional continuous-time generative models, and open up new directions for designing principled, fast, and robust generative systems.

%
%
%
\bibliographystyle{splncs04}
\bibliography{main}

\end{document}